\documentclass[a4paper]{article}

\usepackage{lineno,hyperref}
\modulolinenumbers[5]

\usepackage{graphicx}
\usepackage{amsmath,amsfonts,mathrsfs,amssymb}
\usepackage{bm}
\usepackage{algorithm}
\usepackage[noend]{algorithmic}
\usepackage{hyperref}
\usepackage{subfig}
\usepackage{color}
\usepackage{rotating}
\usepackage{multirow}
\usepackage{placeins}
\usepackage{soul}
\usepackage{marvosym}

\newcommand{\pluseq}{\mathrel{+}=}

\newcommand{\var}{\textit{Var}}

\newcommand{\dataset}{\mathscr{D}}
\newcommand{\datatrain}{\dataset_\text{TRAIN}}

\newcommand{\forest}{\mathcal{E}}
\newcommand{\tree}{\mathcal{T}}
\newcommand{\node}{\mathscr{N}}
\newcommand{\importance}{\mathit{importance}}
\newcommand{\error}{\mathit{e}}
\newcommand{\impurity}{\mathit{impu}}
\newcommand{\prototype}{\mathit{prototype}}
\newcommand{\gini}{\mathit{Gini}}

\newcommand{\sep}{ }

\begin{document}

\author{Matej Petkovi\'{c}\textsuperscript{\Letter}, Dragi Kocev, Bla\v{z} \v{S}krlj, Sa\v{s}o D\v{z}eroski \\
\textit{Jo\v{z}ef Stefan Institute, Ljubljana, Slovenia} \\
\textit{Jo\v{z}ef Stefan International Postgraduate School, Ljubljana, Slovenia}\\
\{matej.petkovic, dragi.kocev, blaz.skrlj, saso.dzeroski\}@ijs.si
}
\date{\today}
\title{Ensemble- and Distance-Based Feature Ranking for Unsupervised  Learning}

\maketitle

\begin{abstract}
In this work, we propose two novel (groups of) methods for unsupervised feature ranking and selection.
The first group includes feature ranking scores (\textsc{Genie3} score, \textsc{RandomForest} score) that are computed from ensembles of predictive clustering trees.
The second method is \textsc{URelief}, the unsupervised extension of the Relief family of feature ranking algorithms.
Using 26 benchmark data sets and 5 baselines, we show that both the \textsc{Genie3} score (computed from the ensemble of extra trees) and the \textsc{URelief} method outperform the existing methods and that Genie3 performs best overall, in terms of predictive power of the top-ranked features.
Additionally, we analyze the influence of the hyper-parameters of the proposed methods on their performance, and show that for the Genie3 score the highest quality is achieved by the most efficient parameter configuration.
Finally, we propose a way of discovering the location of the features in the ranking, which are the most relevant in reality.
\vskip0.1cm
\end{abstract}
feature ranking \sep unsupervised learning \sep tree ensembles \sep extra trees \sep Relief



\section{Introduction}\label{intro}

High-dimensional data are becoming part of everyday life, making the learning of models much harder, slower or even infeasible, unless some dimensionality reduction is performed. 
The problem (high dimensionality) may be solved by \emph{feature ranking} approaches. In the supervised scenario, where the examples in the data are described as $(\bm{x}, y)$ pair, with $\bm{x}$ a tuple of values of descriptive attributes (features) $x_i$ and $y$ a target value, the goal of feature ranking is to assign a real-valued score to each feature, which defines how relevant the feature is for predicting the target $y$. By assessing the relevance correctly, one can i) discard the irrelevant features and make learning predictive models faster/feasible, or ii) understand the data better, which helps when a machine learning expert collaborates with a domain expert.

However, sometimes the examples in the data are hard to label since labeling either demands a lot of manual work (e.g., labeling tweets \cite{petra}), or is expensive for some other reason (e.g., determining the activity of chemical compounds \cite{DiMasi}).

On the spectrum of major machine learning paradigms/settings, semi-supervised learning is situated between supervised learning, which uses only labeled examples (with known values of the targets) to learn predictive models, and unsupervised learning, where only examples without labels are used, typically to group examples into clusters of similar examples. Similarity between examples is determined with respect to a given distance measure, typically Euclidean. Just as we need feature ranking in the context of supervised and semi-supervised learning, we also need it in unsupervised learning. 

Since we do not have targets to predict in unsupervised learning, feature importance in this setting cannot mean how relevant the features are for predicting the targets (and building a predictive model for that purpose). Rather, feature importance in this context measures how relevant a given feature is for constructing a clustering from the given data. Other formulations and motivations of relevance in ranking in an unsupervised context have been considered, such as the one recently proposed by Doquet and Sebag \cite{agnos}: We do not know how relevant is a given feature (for predicting a target), but we can assess how redundant it is in the context of the other features (a feature is redundant if we can predict it accurately from other features). However, we believe the relevance to constructing a clustering captures more appropriately the spirit of unsupervised feature ranking and it is this notion that is embedded within the approaches for unsupervised feature ranking that we put forward in this paper. The intuition of Doquet and Sebag \cite{agnos} corresponds more to feature selection, the task of finding an optimal set of features, where redundancy is to be avoided. Feature selection can be viewed as a special case of feature ranking/ relevance estimation, where every feature is assigned a score of either 1 (the feature is kept) or 0 (the feature is discarded). We can obtain a feature selection by first obtaining a ranking (in terms of relevance scores) and then keeping only the features with the scores exceeding some user-defined threshold \cite{info-fusion:selection-via-ranking}.

In this work, we propose two groups of novel methods for unsupervised feature ranking. Their implementation, as well as the results, are available at \url{https://github.com/Petkomat/unsupervised_ranking}. The first group consists of two ensemble-based feature ranking scores that are computed from ensembles of predictive clustering trees used for hierarchical clustering \cite{pcts}. At each internal node of a tree, the algorithm finds a feature along which the data are most efficiently divided into subclusters. In the leaves, prototypes of the clusters are computed. These prototypes can be seen as a prediction for feature values for a new-coming example, thus the features that are (more often) used in the splits, are (more) relevant for predicting the values of all the features.

The second group consists of only one member named \textsc{URelief}, i.e., the unsupervised extension of the \textsc{Relief} family of algorithms \cite{kira:relief,kononenko:relief}. It does not use any underlying model, but rather estimates the feature relevances via the influence of the features to the distances between the neighboring examples in the data. In other words, if the distance along a given feature strongly contributes to the overall distance between examples, the feature is deemed important. In distance-based clustering, the feature would be important for the clustering of the examples.

The rest of the paper is organized as follows.
Sec.~\ref{sec:related} gives a brief description of related work, in particular the baseline methods that we compare to.
In Sec.~\ref{sec:ensemble-scores}, a detailed description of the proposed ensemble-based scores is given, whereas in Sec.~\ref{sec:URelief-scores} this is done for \textsc{URelief}.
Sec.~\ref{sec:setup} describes the experimental setup, including the evaluation procedure.
Sec.~\ref{sec:results} discusses the experimental results, and Sec.~\ref{sec:conclusions} concludes the paper.

\section{Related Work}\label{sec:related}

While the literature on supervised feature selection has been extensively reviewed and studied \cite{Guyon:JMLR:2003,vir:stanczykJainPregled}, the literature on the topic of unsupervised feature ranking has not been extensively and systematically surveyed. The closest overview is the recent work by Solorio-Fern\'{a}ndez et al. \cite{review:fsel} that overviews methods for unsupervised feature selection. It divides the methods based on the strategy they use for selecting the features, into three groups: filter, wrapper and hybrid. The filter methods evaluate the features based on intrinsic properties of the data, the wrapper methods evaluate the features based on the results of a specific clustering algorithm, and the hybrid methods exploit the advantages of both filter and wrapper methods with a special focus on the balance between the computational cost and the quality of the selected features. Here, we briefly summarize the most prominent and relevant work on the topic of \emph{unsupervised feature ranking}.

To begin with, the \textsc{Scikit-feature} repository \cite{scikit-feature} (\url{https://github.com/jundongl/scikit-feature}) offers unified implementations of some unsupervised feature ranking algorithms. One of them is the Laplace score \cite{laplace}. It is based on the graph theory and the creation of the Laplace matrix of a graph, whose vertices are the examples and whose edges correspond to the level of similarity among the examples.

Next, SPEC (spectral clustering) \cite{spec} is a unification of many existing feature ranking scores. These include the \textsc{Relief} and the Laplace scores.
The actual score that the authors experiment with is thus a result of particular instantiations, such as choosing an appropriate similarity measure, e.g., radial basis kernels.

Furthermore, the similarity matrix (calculated by using a radial-basis kernel) is also one of the building blocks of the MCFS algorithm \cite{mcfs}. However, in addition to calculating this matrix, the algorithm proceeds to solving a generalized eigenvalue problem. It ensures the sparsity of the scores by including a L1-regularization term in the objective function.

Next, another variation of the above approaches is NDFS \cite{ndfs}. Here, L2,1-regularization is applied rather than L1-regularization. The best features are those which are the most closely related to the labels of the clusters constructed within the algorithm.

The last method that we discuss here is \textsc{AgnoS-S}. This is the best performing variation among the unsupervised ranking methods proposed by Doquet and Sebag \cite{agnos} in their award-wining paper at the ECML 2019 conference. This is an auto-encoder-based feature ranking method, where slack variables are included into the auto-encoder (and the loss function) to ensure that more important features get higher weights. The method also estimates the intrinsic dimension of the data \cite{intrinsic}, determining the width of the middle layer of the auto-encoder, which the method optimizes.

We use the above approaches as baselines when evaluating the quality of the methods we introduce in this paper.

\section{Ensemble-Based Feature Ranking}\label{sec:ensemble-scores}
The first two proposed feature ranking scores are embedded in ensembles of predictive-clustering trees (PCTs). 
In this section, we first describe the PCTs, proceed to ensembles thereof, and conclude with feature ranking scores.

\subsection{Single PCTs}
PCTs are a fully modular generalization of standard decision trees \cite{tdidt} where three groups of attributes need to be defined.

The attributes that can be used in the tests (splits) in the internal nodes of the trees (line \ref{alg:btest1} in Alg.~\ref{alg:bestTest}) are \emph{descriptive} attributes (or features). The attributes that are used to assess the quality of split candidates, i.e., are part of the formula of impurity function $\impurity{}$, are \emph{clustering} attributes. The attributes that are predicted in the leaves, i.e., returned by the $\prototype{}$ function, are \emph{target} attributes.

When PCTs are used in predictive modeling, clustering and target attributes mostly coincide (semi-supervised learning is one of the exceptions \cite{jurica:phd}), and there is typically only one such attribute (multi-target tasks are an exception \cite{Kocev:Journal:2013}). For example, in classification, we are predicting the value of a single nominal attribute $y$, which is also used in impurity calculations. The remaining attributes are descriptive.

However, we use PCTs for (hierarchical) clustering. Here, the groups of descriptive, clustering, and target attributes coincide, since any of the attributes can be used in a test, all of them are taken into account when computing the impurity of clusters, and all of them are "predicted" when the prototypes of the clusters are computed in the leaves.

The exact pseudocode for inducing a PCT is given in Alg.~\ref{alg:pct}, which takes as its input a subset of examples $E\subseteq \datatrain{}$, and returns a PCT.
\vskip-1em
\noindent\begin{minipage}[!T]{0.44\textwidth}
    \begin{algorithm}[H]
    \caption{PCT($E$)}\label{alg:pct}
     \begin{algorithmic}[1]
         \STATE $(t^*,h^*,\mathcal{P}^*) = \mathrm{BestTest}(E)$\label{alg:pct:search}
         \IF{$t^* = \mathit{none}$}
            \STATE \textbf{return} $\mathit{Leaf}(\prototype{}(E))$\label{alg:pct:leaf}
        \ELSE
             \FOR{\textbf{each} $E_i \in \mathcal{P^*}$}
                \STATE $\mathit{tree}_i$ = PCT($E_i$)
             \ENDFOR
             \STATE \textbf{return} $\mathit{Node}(t^*,\;\bigcup_i \{\mathit{tree}_i\})$\label{alg:pct:internal}
         \ENDIF
    \end{algorithmic}
    \end{algorithm}
\end{minipage}
\begin{minipage}[!T]{0.59\textwidth}
\begin{algorithm}[H] 
    \caption{$\mathrm{BestTest}(E)$}\label{alg:bestTest}
 \begin{algorithmic}[1]
     \STATE{$(t^*,h^*,\mathcal{P}^*) = (\mathit{none},0,\emptyset)$}
     \FOR{\textbf{each} test $t$} \label{alg:btest1}
         \STATE{$\mathcal{P} = $ partition induced by $t$ on $E$}\label{alg:partition}
         \STATE{$h = |E|\impurity{}(E) -\sum_{E_i \in \mathcal{P}} |E_i| \impurity{}(E_i)$}\label{alg:heurCom}
        \IF{$h > h^*$} \label{alg:icv}
            \STATE $(t^*,h^*,\mathcal{P}^*) = (t,h,\mathcal{P})$ \label{alg:btest2}
        \ENDIF
     \ENDFOR
     \STATE \textbf{return} $(t^*,h^*,\mathcal{P}^*)$
 \end{algorithmic}
 \end{algorithm}
\end{minipage}
\vskip1em
\noindent In clustering (unsupervised learning), the impurity function for a given subset $E\subseteq \datatrain{}$ is defined as the average of $\impurity{}(E) = \frac{1}{n}\sum_{i= 1}^n \impurity{}(E, x_i)$ of the clustering-attribute impurities $\impurity{}(E, x_i)$, which are defined as follows.

For nominal variables $x$, the impurity is defined in terms of the Gini Index $\gini{}(E, x) = 1 - \sum_v p_E^2(v)$, where the sum goes over the possible values $v$ of the variable $x$, and $p_E(x)$ is the relative frequency of the value $v$ in the subset $E$. To avoid any bias towards variables with high overall impurity, the impurity is defined as the normalized Gini value, i.e., 
$\impurity{}(E, x) = \gini{}(E, x) / \gini{}(\datatrain{}, x)$.

For numeric variables $x$, the impurity is similarly defined as the normalized variance of $x$, i.e., $\impurity{}(E, x) = \var{}(E, x) / \var{}(\datatrain{}, x)$.

The prototype function returns a vector $[\hat{x}_1, \dots, \hat{x}_n]$, where $\hat{x}_i$ is a mean of the attribute $x_i$ if the attribute is numeric, and the mode of the attribute, if the attribute is nominal. The means and modes are computed from the training examples in a given leaf.

\subsection{Ensembles of PCTs}
Single trees are known for their instability, i.e., undesirably high variance \cite{Breiman01a:jrnl}. To overcome that, ensemble methods such as bagging \cite{bagging}, random forests \cite{Breiman01a:jrnl} and ensembles of extremely randomized trees  (extra trees) \cite{geurts:extraT} have been proposed. All of these methods have been adapted to learn ensembles of PCTs \cite{Kocev:Journal:2013,Kocev:Journal:2020}.

A tree ensemble is a set of trees which are grown on a possibly manipulated training set and/or a randomized tree induction algorithm. More precisely, in {\bf bagging}, every tree in an ensemble is grown on an independent bootstrap replicate of the training set $\datatrain{}$. {\bf Random forests} additionally consider only a random subset of descriptive attributes, when finding the best test (line \ref{alg:btest1} of Alg.~\ref{alg:bestTest}).
The subset is drawn independently in every internal node.

{\bf Extra trees}, as the name suggests, go even further regarding randomization and, as the random forests, first choose a subset of descriptive attributes, and then evaluate only one test per attribute. For example, if $x_i$ is one of the chosen numeric features, random forests evaluate the tests $x_i\leq \vartheta$, for all possible values of $\vartheta$, whereas extra trees additionally randomly choose only one $\vartheta{}_i$, for each chosen feature $x_i$.

Originally, extra trees do not use bootstrapping. However, we have included it in learning extra PCTs since i) our preliminary experiments show that it is beneficial to use it (especially when there are many binary features in the data), and ii) one of the feature scores described in the next section requires bootstrapping.

\subsection{Feature Ranking Scores}
We propose to use two feature ranking scores that can be computed from any of the ensembles described in the previous section.

The first one is \textsc{Genie3} \cite{genie3}, originally defined for predictive modeling tasks, such as classification or regression. Within the framework of predictive clustering, we can extend the score to unsupervised modeling as
\begin{equation}
\label{eq:genie}
\importance_\textsc{Genie3}(x_i) = \frac{1}{|\forest|} \sum_{\tree \in \forest} \sum_{\node \in \tree(x_i)} h^*(\node)\text{,}\\
\end{equation}
where $\forest{}$ is an ensemble of trees $\tree{}$, and $\tree{}(x_i)$ is the set of internal nodes of the tree $\tree{}$ where $x_i$ is part of the test. Finally, $h^*(\node)$ is the quality of the test in the node $\node{}$, as assessed by the heuristic in Alg.~\ref{alg:bestTest}, line \ref{alg:heurCom}.

Note that $h^*(\node{})$ is proportional to the number of examples that reach the node $\node{}$ (since those nodes have bigger influence), and the impurity reduction value. 
One could thus also use a simpler version of the \textsc{Genie3} score (namely Symbolic score \cite{petkomat:mtr}) that ignores the impurity reduction factor, and weights every appearance of the feature $x_i$ by the number of examples that reach that node.

The second feature ranking score that we propose is a generalization of the \textsc{RandomForest} score \cite{Breiman01a:jrnl}. Instead of evaluating a feature directly via the quality of the tests, this approach compares how much introducing some noise in the feature degrades the performance of a model. More precisely, for every tree $\tree{}$ in the ensemble, we first define the set of out-of-bag examples $\text{OOB}_\tree{}$, i.e., the examples that were not chosen into the bootstrap replicate of the training set. Next, we define the set $\text{OOB}_\tree{}^i$ that is obtained from the set $\text{OOB}_\tree{}$ by permuting the values of the feature $x_i$.
Let $\error{}$ be some error measure (e.g., the average of the per-feature variances in a cluster). Then, the \textsc{RandomForest} (RF) score is defined as
\begin{equation}
    \label{eq:rf}
\importance_\text{RF}(x_i) = \frac{1}{|\forest|}\sum_{\tree \in \forest}
\frac{\error(\text{OOB}_\tree^i) - \error(\text{OOB}_\tree)}{\error(\text{OOB}_\tree)}\text{,}
\end{equation}
i.e., as the relative increase of the error measure $\error{}$. The larger the increase, the more important the feature. The score has been originally defined for random forests (hence the name), but can be used with practically any ensemble approach based on bagging.

\subsection{Time complexity}\label{sec:times-ensemble}
Since the sets of descriptive, clustering and target attributes coincide with the number of all attributes, we denote their sizes by $n$. The other dimension of the data set, i.e., the number of examples, is denoted by $m$.

In the analysis, we assume that the trees are approximately balanced, i.e., their depth is $\mathcal{O}(\log m)$. The largest amount of computation is performed when estimating the quality of the tests. In a given node, evaluating split candidates that a single attribute yields, requires $\mathcal{O}(m'\log m' + m' n)$ steps (sorting the $m'$ examples that reach the node according to their attribute values) and efficiently evaluating either one (extra trees) or all thresholded tests (bagging, random forests).

Taking into account that there are $n'$ chosen features in each node (for bagging, $n' = n$), and the assumed depth of the tree, we can derive the total time complexity of growing a tree: $\mathcal{O}((m\log m + m n)n' \log m) = \mathcal{O}((\log m + n)n' m\log m)$. Thus, bagging is quadratic in the number of features which is a problem if trees are not grown in parallel. The latter is also the reason why we did use an unsupervised adaptation of gradient-boosted trees \cite{xgb}. One can loose the inner $\log m$ term by presorting the data with respect to all the features in advance.

Since the trees are grown independently, the algorithm is easy to parallelize, and the time complexity of growing any of the ensembles is the same.

After that, the cost of computing the \textsc{Genie3} ranking is negligible, since the ensemble stores the quality of the nodes and we can traverse the nodes of a single tree in time $\mathcal{O}(m)$. Computing the \textsc{RandomForest} ranking is a bit more time-consuming. We have to permute the values of a feature, and send the examples trough each tree, for every feature. This is done in $\mathcal{O}(n m\log m)$ time. This is still faster than inducing a tree.

\section{{\normalsize UR}{\scriptsize ELIEF}-based Feature Ranking}\label{sec:URelief-scores}
In contrast to ensemble-based feature ranking scores, the \textsc{Relief} family of feature ranking algorithms does not use any predictive model.
Its members can handle various predictive modeling tasks, including classification \cite{kira:relief}, regression \cite{kononenko:relief}, and others \cite{petkomat:mtr,petkomat:mlc:relief,reyes}.

In this work, we extend the \textsc{Relief} family to unsupervised ranking, by again taking the predictive-clustering point of view. We follow the main intuition behind \textsc{Relief} \cite{kononenko:relief}, which for two examples that are close to each other (in the clustering space) states: The feature $x_i$ is relevant if the differences in the target space between the two examples are notable if and only if the differences in the feature values of $x_i$ between these two examples are notable.

As in PCT induction above, we again define clustering and target space to consist of all $n$ attributes. We estimate the relevance of the descriptive attributes, which again comprise all $n$ attributes.

Being close or having a notable difference is operationalized in the algorithm  via the distances in the appropriate spaces. For the clustering (target) space $\mathcal{X}$ spanned by the domains $\mathcal{X}_i$ of the features $x_i$, we have
\begin{equation}
\label{eqn:metric}
d_\mathcal{X}(\bm{x}^1, \bm{x}^2) = \frac{1}{n}\sum_{i = 1}^n d_i(\bm{x}^1, \bm{x}^2);\;\;
d_i(\bm{x}^1, \bm{x}^2)=
\begin{cases}
\;\bm{1}[\bm{x}_i^1 \neq \bm{x}_i^2]&: \mathcal{X}_i \nsubseteq \mathbb{R}\\
\frac{|\bm{x}_i^1 -\bm{x}_i^2|}{\max\limits_{\bm{x}} \bm{x}_i - \min\limits_{\bm{x}}\bm{x}_i} &: \mathcal{X}_i\subseteq \mathbb{R}
\end{cases}
\end{equation}
where $\bm{1}$ denotes the indicator function. These distances are then used in the estimation of the expression
\begin{equation}\label{eqn:URelief}
      P(\bm{x}_i^1 \neq \bm{x}_i^2 \mid  \bm{x}^1 \neq \bm{x}^2) - P(\bm{x}_i^1 \neq \bm{x}_i^2 \mid  \bm{x}^1 = \bm{x}^2)\text{,}
\end{equation}
which is the output of the \textsc{URelief} algorithm. Above, $\bm{x}^{1, 2}\in\mathcal{X}$ are two examples,
and $\bm{x}_i^{1, 2}$ values of the feature $x_i$ for these two examples. Clearly, if the features are numeric, the conditional probabilities above may be trivial or not be well defined. As explained below, they serve merely as a motivation for the algorithm.

The unsupervised version of the algorithm, \textsc{URelief} is shown in Alg.~\ref{alg:URelief}.
Within the code, the probability $P_{\text{diffAttr}, i} = P(\bm{x}_i^1 \neq \bm{x}_i^2)$ is modeled as the distance $d_i$ (line \ref{alg:URelief:pA}), whereas the probability $P_{\text{diffClus}} = P(\bm{x}^1 \neq \bm{x}^2)$ is modeled as the distance $d_\mathcal{X}$ (line \ref{alg:URelief:pC}). The conditional probabilities are estimated via the Bayes formula (line \ref{alg:URelief:w}). The product of two events, e.g., $P(\bm{x}_i^1 \neq \bm{x}_i^2 \;\land\;  \bm{x}^1 \neq \bm{x}^2)$ is modeled as the product of the two probabilities: This is used together with the Bayes formula in line \ref{alg:URelief:w}.
\begin{algorithm}[t] 
	\caption{URelief($\datatrain$, $I$, $K$)}
	\label{alg:URelief}
	\begin{algorithmic}[1]
		\STATE{$\bm{w}$, $\bm{P}_\text{diffAttr, diffClus}$, $\bm{P}_\text{diffAttr} = $ zero lists of length $n$}
		\STATE{$P_\text{diffTarget}= 0.0$}
		\FOR{iteration $= 1, 2,\dots, I$}
			\STATE{$\bm{r} = $ random example from $\datatrain{}$}\label{line:rndEx}
			\STATE{$\bm{n}_1, \bm{n}_2, \dots, \bm{n}_K =$ $K$ nearest neighbors of $\bm{r}$ with respect to $d_\mathcal{X}$}\label{line:knn}
			\FOR{$k = 1,2, \dots, K $}
				\STATE{$P_\text{diffClus} \pluseq d_\mathcal{X}\big(\bm{r}, \bm{n}_k\big) / (I K)$}\label{alg:URelief:pC}
				\FOR{$i = 1, 2, \dots, n$}\label{line:updateStart}
					\STATE{$\bm{P}_\text{diffAttr}[i] \pluseq d_i\big(\bm{r}, \bm{n}_k\big) / (I K)$}\label{alg:URelief:pA}
					\STATE{$\bm{P}_\text{diffAttr, diffClus}[i] \pluseq d_i\big(\bm{r}, \bm{n}_k\big) d_\mathcal{X}\big(\bm{r}, \bm{n}_k\big) / (I K)$}\label{alg:URelief:pAC}
				\ENDFOR
			\ENDFOR
		\ENDFOR
		\FOR{$i = 1,2, \dots, n$}
			\STATE{$w_i = \frac{\bm{P}_\text{diffAttr, diffClus}[i]}{P_\text{diffClus}}
			- \frac{\bm{P}_\text{diffAttr}[i] - \bm{P}_\text{diffAttr, diffClus}[i]}{1 - P_\text{diffClus}}$}\label{alg:URelief:w}
		\ENDFOR
		\RETURN{$\bm{w}$}
	\end{algorithmic}  
\end{algorithm}

The algorithm is iterative. On every iteration, an example $\bm{r}$ is chosen from the training set (line \ref{line:rndEx}), and its nearest neighbors $\bm{n}_k$ are computed in the clustering space. After that, the estimates for the needed probabilities are updated according to the distances between $\bm{r}$ and $\bm{n}_k$. The updates are normalized, so that the final estimate of every probability falls into the interval $[0, 1]$. Finally, in line \ref{alg:URelief:w}, the weights (feature importance scores) are computed.

The \textbf{time complexity} of \textsc{URelief} can be estimated as follows. 
To make full use of the efficient vectorized implementation of scipy's \cite{scipy} \texttt{cdist} for computing the distances among the examples, we first compute all nearest neighbors. This takes $\mathcal{O}(I m n)$ steps. Since the number of iterations is typically set to be a proportion of $m$, the number of steps is $\mathcal{O}(m^2 n)$.
When updating the probability estimates, only the $K$ nearest neighbors are considered, so the total number of steps is also $\mathcal{O}(m^2 n)$, since $K$ is upper-bounded by some constant.

Thus, regarding the time complexity, \textsc{URelief} is more appropriate for high-dimensional data than ensemble-based feature rankings, since it is linear in the number of features.
However, the results show that this comes at a performance price.

\section{Experimental Setup}\label{sec:setup}
In this section, we describe the experimental setup of the empirical evaluation of the proposed methods. We state the experimental questions and competing methods, describe the evaluation procedure, give a brief description of the data sets, and finish with the parameter instantiation of the different ranking methods.

The evaluation is based on the following experimental questions:
\begin{enumerate}
    \item Which of the proposed ensemble scores is better?
    \item Are the proposed feature ranking scores state-of-the-art (SOTA)?
    \item What is the influence of the parameters of the proposed methods on the quality of the produced rankings?
    \item How to locate the relevant features in a ranking?
\end{enumerate}

To answer the first question, we compare the performance of the two ensemble-based ranking scores with Friedman's (statistical) test \cite{demsar}. The evaluation procedure is discussed in Section~\ref{sec:eval_procedure}.

To answer the second question, \textsc{URelief}, the better of the two ensemble-based scores, and five current SOTA methods are compared via Friedman's test,
together with Nemenyi's post-hoc test \cite{demsar}. The considered competitors are \textsc{AgnoS-S}, the Laplace score, NDFS, MCFS and SPEC, all briefly described in Sec.~\ref{sec:related}.

To answer the third question, we compare the performances of the proposed scores obtained with different parameter configurations. We vary the ensemble method and the subset size for ensemble-based scores. We vary the numbers of neighbors and iterations for \textsc{URelief}.

To answer the fourth question, we construct a curve that measures the performance of the sets (of different sizes) of top-ranked features.
A detailed description of the approach taken is given in Sec.~\ref{sec:curves}.

\subsection{Datasets}\label{sec:data}
We first planned to use all the 29 available datasets from the \textsc{Scikit-feature} repository. To assure that the considered datasets are independent, as the statistical tests assume, we had to exclude three of them: \texttt{lung\_small} (but \texttt{lung} is present), \texttt{orlraws10P} (but \texttt{ORL} is present) and \texttt{warpAR10P} (but \texttt{warpPIE10P} is present). Those included - together with the number of features and examples, are listed in Tab.~\ref{tab:data} and available for download at \url{https://github.com/jundongl/scikit-feature}. Note that they typically have a low number of examples and a high number of features.

\begin{table}[htbp]
  \centering
  \caption{The basic characteristics of the considered data sets.}
    \begin{tabular}{lrrl}
    \hline
    data set & \multicolumn{1}{l}{examples} & \multicolumn{1}{l}{features} & domain \\
    \hline
    ALLAML & 72    & 7129 & biology \\
    arcene & 200   & 10000 & mass spectrometry \\
    BASEHOCK & 1993  & 4862 & text data \\
    Carcinom & 174   & 9182& biology \\
    CLL-SUB-111 & 111   & 11340& biology \\
    COIL20 & 1440  & 1024 & face image\\
    colon & 62    & 2000 & biology\\
    gisette & 7000  & 5000 & digit recognition \\
    GLI-85 & 85    & 22283& biology \\
    GLIOMA & 50    & 4434& biology \\
    Isolet & 1560  & 617 & spoken letter recognition\\
    leukemia & 72    & 7070& biology \\
    lung  & 203   & 3312& biology \\
    lymphoma & 96    & 4026& biology \\
    madelon & 2600  & 500 & artificial \\
    nci9  & 60    & 9712& biology \\
    ORL   & 400   & 1024& face image \\
    PCMAC & 1943  & 3289 & text data\\
    pixraw10P & 100   & 10000& face image \\
    Prostate-GE & 102   & 5966& biology \\
    RELATHE & 1427  & 4322& text data \\
    SMK-CAN-187 & 187   & 19993& biology \\
    TOX-171 & 171   & 5748& biology \\
    USPS  & 9298  & 256 & hand written image \\
    warpPIE10P & 210   & 2420& face image \\
    Yale  & 165   & 1024& face image \\
    \hline
    \end{tabular}%
  \label{tab:data}%
\end{table}%

\subsection{Evaluation Procedure}\label{sec:eval_procedure}
In the supervised learning setting, one of the popular procedures for evaluating feature ranking algorithms is the use of the weighted $k$-nearest neighbors ($k$NN) algorithm \cite{knn-wettschereck} in which the distance among the examples employed weights features by their importance and is defined as $$d(\bm{x^1}, \bm{x^2}) = \left(\sum_{i = 1}^n \importance{}(x_i) (\bm{x^1}_i - \bm{x^2}_i)^p\right)^{1 / p}$$ 
for some $p > 0$. For example, $p = 2$ results in the weighted Euclidean distance, and the better the performance, the better are the feature-importance estimates \cite{knnweights}. In the unsupervised scenario, choosing the correct evaluation procedure is trickier.

For example, the above procedure cannot be extended to the unsupervised scenario, since we would predict the same attributes that are used for the distance computation.
As a consequence, if one, for example, measures the performance of the model in terms of mean squared error, it can be proven 
that the performance of the nearest-neighbors model is optimized when the weights $\importance{}(x_i)$ are all equal.

Another popular (yet sometimes not appropriate procedure) for evaluating unsupervised feature rankings is to i) use a supervised data set, ii) compute a ranking ignoring the target variable, iii) take some top-ranked features and learn a model that predicts the target variable. This approach only makes sense when the so-called clustering hypothesis holds \cite{ssl-intro}, i.e., when \emph{clusters of data examples (as computed in the descriptive space) well resemble the distribution of target values}.
Since the considered datasets also contain the target variable, they allow us to check this hypothesis by computing the Adjusted Rand Index (ARI) \cite{ari} between the actual class labels and the labels obtained from the $k$-means clustering, where $k$ was set to the number of classes in a given data set.

\begin{figure}
    \centering
    \includegraphics[scale=0.5]{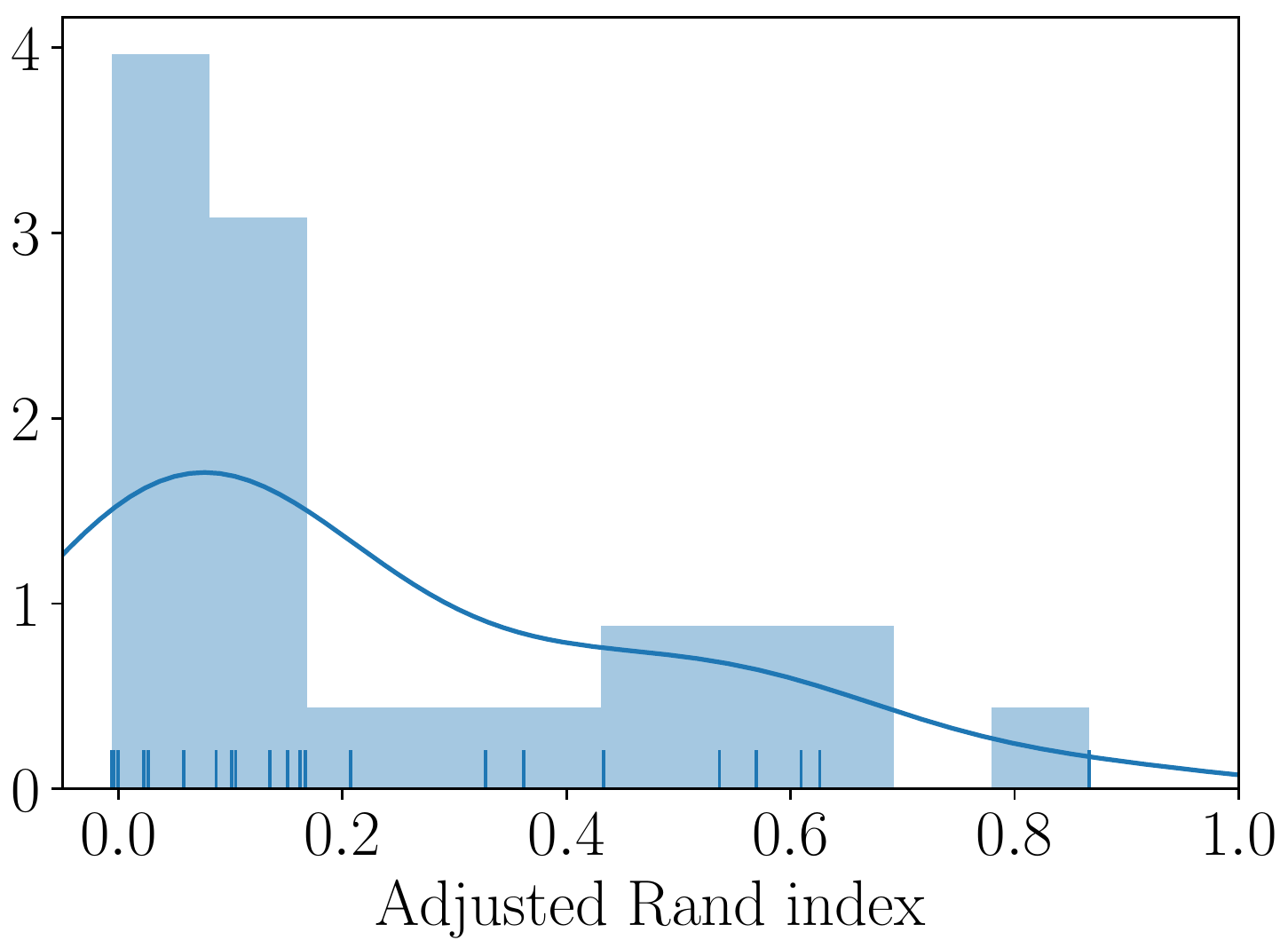}
    \caption{The distribution of the ARI values (as estimated by kernel-density estimation in Python's module seaborn (Available at \url{https://seaborn.pydata.org/}) over the considered data sets. The ARI value for a given data set (represented as a single tick) is the median of the ARI values over ten runs of the k-means algorithm, where k equals the number of classes in a given data set.
    }
    \label{fig:ari}
\end{figure}

The ARI can take the values from the interval $[-1, 1]$, and is adjusted so that ARI = 0 if the two labelings are independent, i.e., the clusters do not tell us anything about the target. The distribution of the ARI scores over the considered data sets is given in Fig.~\ref{fig:ari} and shows that there are many data sets with ARI close to 0. Therefore, the supervised approach to evaluating unsupervised feature rankings is also inappropriate.

Therefore, we adopt another widely-used approach \cite{agnos} where some fixed number of features is chosen, and a predictive model is built using only this many top-ranked features. We build the models that use top 1, 2, 4, 8, 16, etc. features to analyze a given feature ranking in-depth (see Sec.~\ref{sec:curves}). Therefore, we choose 16 features when comparing the qualities of feature rankings, since 16 is the closest to the 20 features that Doquet and Sebag \cite{agnos} use. When choosing a model, we again follow Doquet and Sebag \cite{agnos}, and use a nearest-neighbors regressor. However, since some of the data sets contain a rather low number of examples (e.g., 50), we set the number of neighbors to $1$.

Note that the evaluation procedure of Doquet and Sebag \cite{agnos} is not entirely correct, since it uses the same data for learning the ranking and evaluating it, and therefore measures only how well the ranking \emph{overfits} the data.
We fix this by using 10-fold cross-validation: A feature ranking is learned from a training fold, and the $1$NN regressor with data from the training fold is tested on the testing fold. As the evaluation measure, we use mean squared error: All the features in the data are numeric, since some competing methods cannot handle nominal features.

\subsection{Parameter instantiation}
Previous experiments with the semi-supervised version of \textsc{Relief} \cite{petkomat:ssl} showed that the optimal number of neighbors is around 30, so we set $K = 30$ in \textsc{URelief}. 
Regarding the number of iterations $I$, higher values means more accurate estimates, so we use $I = m$, $m$ being the number of examples in the data.
Since the number of examples is not very high, this value of $I$ is not prohibitively large. We experiment with the other parameter configurations in Sec.~\ref{sec:hyperparameters}, considering different numbers of neighbors and iterations.

For calculating the ensemble-based rankings, we set the size of the subset of chosen features to $n' = \log_2 n$ (rounded up), as suggested by Breiman \cite{Breiman01a:jrnl}. Moreover, this number is sufficiently low to make tree induction time-efficient.
We use the extra tree ensembles (they evaluate only $n'$ tests per internal node)
and induce the trees in parallel, using the Slovenian national supercomputing network. The number of trees was set to 100 and they were fully grown.
We also experiment with other parameter configurations from Section~\ref{sec:hyperparameters}, considering different subset sizes and ensemble methods.

For the baselines implemented in the \textsc{Scikit-feature} repository, we use the default settings, following Doquet and Sebag \cite{agnos}.
If a method demands the number of clusters at input (MCFS, NDFS), we set it to the number of classes in the given data set.
For \textsc{AgnoS-S}, we use the shape of the network and the training parameters as reported by Doquet and Sebag \cite{agnos}.

\section{Results}\label{sec:results}

Below we discuss the results of the evaluation in light of our experimental questions.

\subsection{Which Ensemble Score is Better?}

To answer this question, we directly compare the cross-validated MSE performance of the $1$NN regressors that use the top-ranked features,
as determined by the \textsc{Genie3} and \textsc{RandomForest} scores calculated from extra tree ensembles. Their performance is reported in Table~\ref{tab:genie-vs-URelief}, together with the corresponding average rank: for a given data set, the better score gets rank 1, and the other gets rank 2.

\begin{table}[htbp]
  \centering
  \caption{The performance (in terms of MSE of the corresponding $1$NN regressors using the 16 top-ranked features) of \textsc{Genie3} and \textsc{RandomForest} scores computed from ensembles of extra trees. Additionally, the average ranks of both methods are reported.}
    \begin{tabular}{lrr}
\hline
\multicolumn{1}{l}{Dataset} & \multicolumn{1}{l}{Genie3} & \multicolumn{1}{l}{Random Forest} \\
    \hline
    ALLAML & 1.15  & \textbf{1.14} \\
    arcene & 40.92 & \textbf{40.79} \\
    BASEHOCK & \textbf{0.17}  & 0.18 \\
    Carcinom & \textbf{0.33}  & 0.34 \\
    CLL-SUB-111 & \textbf{1726.00} & 2449.00 \\
    COIL20 & 0.12  & \textbf{0.10} \\
    colon & 1.43  & \textbf{1.42} \\
    gisette & \textbf{269.00} & 275.00 \\
    GLI-85 & 1516.00 & \textbf{1508.00} \\
    GLIOMA & \textbf{0.23}  & 0.24 \\
    Isolet & 0.37  & \textbf{0.36} \\
    leukemia & \textbf{1.84}  & 1.87 \\
    lung  & 0.29  & 0.29 \\
    lymphoma & \textbf{1.73}  & 1.87 \\
    madelon & \textbf{31.79} & 33.61 \\
    nci9  & \textbf{1.62}  & 1.66 \\
    ORL   & \textbf{24.90} & 34.20 \\
    PCMAC & \textbf{0.18}  & 0.19 \\
    pixraw10P & 6.65  & \textbf{6.19} \\
    Prostate-GE & 0.22  & \textbf{0.21} \\
    RELATHE & \textbf{0.22}  & 0.23 \\
    SMK-CAN-187 & 0.34  & 0.34 \\
    TOX-171 & \textbf{206.00} & 217.00 \\
    USPS  & \textbf{0.33}  & 0.48 \\
    warpPIE10P & \textbf{16.90} & 22.30 \\
    Yale  & \textbf{46.70} & 51.80 \\
    \hline
    Average Rank   & \textbf{1.31}  & 1.69 \\
    \hline
    \end{tabular}%
  \label{tab:genie-vs-URelief}%
\end{table}%

Results suggest that \textsc{Genie3} feature rankings are, on average, better than \textsc{RandomForest} ones. The differences in performance between the two methods are also statistically significant (at the significance level of $0.05$): The $p$-value from the Friedman's test is $0.0476$. This is why we will use the \textsc{Genie3} feature importance score in the main comparison of the methods presented below.

\subsection{Are the Proposed Scores SOTA?}
To answer this question, we compare the performance of \textsc{Genie3} rankings, \textsc{URelief} rankings, and the baseline rankings (or, rather, the performance of the corresponding $1$NN regressors), on different data sets. The corresponding MSE values are reported in Table~\ref{tab:sota}.

\begin{table}[htbp]
  \scriptsize
  \centering
  \caption{The performance of the $1$NN regressors that correspond to the \textsc{Genie3}, \textsc{URelief} and the baseline rankings, for every data set. The best performance per dataset is given in bold typeface. 
  Additionally, the average rank for each of the feature ranking methods is given at the bottom.}
    \begin{tabular}{lrrrrrrr}
\hline
\multicolumn{1}{l}{Dataset} & \multicolumn{1}{c}{\textsc{Genie3}} & \multicolumn{1}{c}{\textsc{URelief}} & \multicolumn{1}{c}{Laplace} & \multicolumn{1}{c}{SPEC} & \multicolumn{1}{c}{MCFS} & \multicolumn{1}{c}{NDFS} & \multicolumn{1}{c}{\textsc{AgnoS-S}} \\
    \hline
    ALLAML & 1.15  & \textbf{1.11}  & 1.16  & 1.15  & 1.16  & 1.15  & 1.22 \\
    arcene & \textbf{40.90} & 51.30 & 42.40 & 76.20 & 42.40 & 42.40 & 80.60 \\
    BASEHOCK & 0.17  & 0.18  & 0.18  & 0.18  & 0.18  & 0.18  & \textbf{0.16} \\
    Carcinom & \textbf{0.33}  & \textbf{0.33}  & 0.36  & 0.36  & 0.35  & 0.36  & 0.36 \\
    CLL-SUB-111 & \textbf{1726.00} & 1823.00 & 1735.00 & 2350.00 & 2429.00 & 1744.00 & 2374.00 \\
    COIL20 & 0.12  & 0.11  & 0.16  & 0.23  & 0.28  & \textbf{0.09}  & 0.18 \\
    colon & \textbf{1.43}  & 1.49  & 1.50  & 1.50  & 1.50  & 1.50  & 1.49 \\
    gisette & \textbf{269} & 278.00 & 291.00 & 315.00 & 291.00 & 291.00 & 304.00 \\
    GLI-85 & \textbf{1516.00} & 1759.00 & 1521.00 & 1783.00 & 1558.00 & 1655.00 & 1712.00 \\
    GLIOMA & \textbf{0.23}  & \textbf{0.23}  & 0.24  & 0.27  & \textbf{0.23}  & 0.24  & 0.27 \\
    Isolet & 0.37  & 0.48  & 0.37  & 0.45  & 0.42  & 0.38  & \textbf{0.36} \\
    leukemia & 1.84  & \textbf{1.83}  & 1.91  & 1.90  & 1.91  & 1.91  & 1.94 \\
    lung  & \textbf{0.29}  & \textbf{0.29}  & 0.30  & 0.33  & \textbf{0.29}  & 0.31  & 0.31 \\
    lymphoma & \textbf{1.73}  & 1.81  & 2.01  & 2.01  & 2.01  & 2.01  & 1.85 \\
    madelon & 31.79 & 31.83 & 33.62 & 33.62 & 33.62 & 33.62 & \textbf{34.61} \\
    nci9  & 1.62  & 1.64  & \textbf{1.60}  & 1.66  & \textbf{1.60}  & \textbf{1.60}  & 1.76 \\
    ORL   & \textbf{24.98} & 27.60 & 34.19 & 34.19 & 34.19 & 34.19 & 23.19 \\
    PCMAC & 0.18  & 0.19  & \textbf{0.16}  & 0.18  & 0.18  & 0.19  & \textbf{0.16} \\
    pixraw10P & 6.65  & 7.93  & 9.30  & 8.48  & 9.30  & 9.30  & \textbf{6.48} \\
    Prostate-GE & \textbf{0.22}  & 0.23  & \textbf{0.22}  & 0.30  & 0.23  & \textbf{0.22}  & 0.24 \\
    RELATHE & \textbf{0.22}  & \textbf{0.22}  & 0.23  & 0.24  & 0.23  & 0.23  & 0.32 \\
    SMK-CAN-187 & \textbf{0.34}  & 0.35  & 0.35  & \textbf{0.34}  & 0.35  & 0.35  & 0.38 \\
    TOX-171 & \textbf{206.00} & 212.00 & 207.00 & 221.00 & 237.00 & 219.00 & 209.00 \\
    USPS  & 0.33  & 0.35  & 0.33  & 0.38  & 0.40  & \textbf{0.29}  & 0.34 \\
    warpPIE10P & 16.97 & 16.45 & 30.28 & 30.28 & 30.28 & 30.28 & \textbf{15.30} \\
    Yale  & 46.77 & 51.29 & 56.76 & 56.76 & 56.76 & 56.76 & \textbf{42.73} \\
    \hline
    Avevare rank   & \textbf{1.90}  & 3.42  & 4.04  & 5.25  & 4.60  & 4.46  & 4.33 \\
    \hline
    \end{tabular}%
  \label{tab:sota}%
\end{table}%

\begin{figure}
    \centering
    \includegraphics[scale=.6]{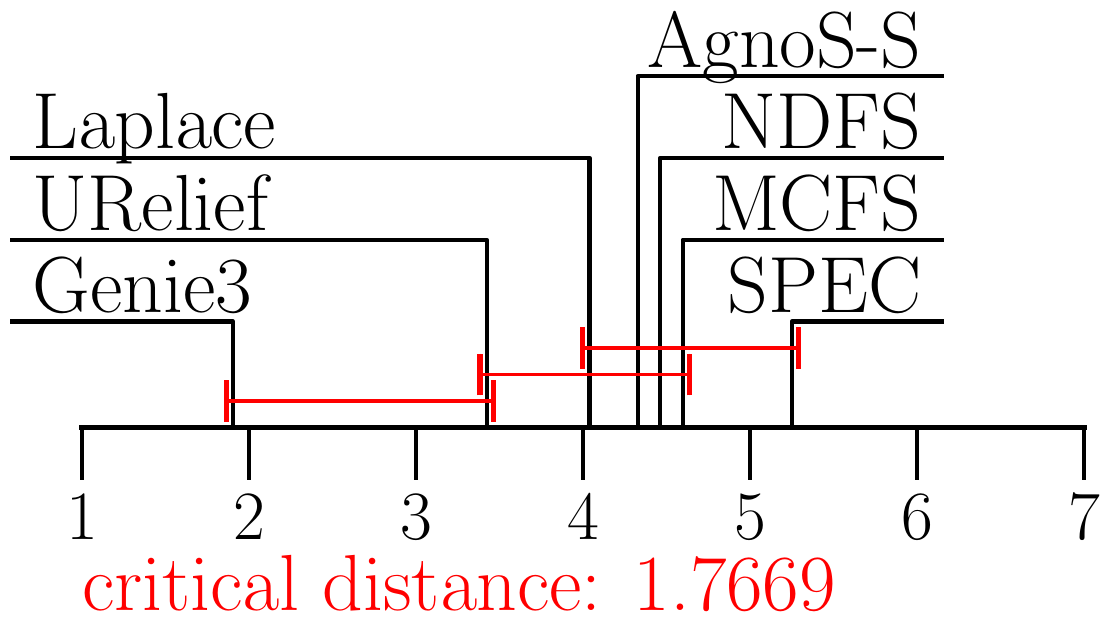}
    \caption{Average rank diagram.}
    \label{fig:sota}
\end{figure}

Comparing the methods we propose (\textsc{Genie3, URelief}) to their competitors in terms of average ranks, cf.~Table~\ref{tab:genie-vs-URelief}, the best-performing feature-ranking method is \textsc{Genie3}. \textsc{URelief} also performs well, as
it has the second-best average rank. To investigate whether the differences in performance among the algorithms are statistically significant,
we apply Friedman's statistical test (since we compare more than two methods). The null hypothesis is rejected at the significance level of $0.05$,
since the $p$-value equals $9.4\cdot 10^{-8}$. Thus, we can proceed to the Nemenyi's post-hoc test that reveals where the differences occur.
The results of these tests are presented with the average rank diagram in Figure~\ref{fig:sota}.

In addition to the average ranks of the algorithms (the lower, the better), the critical distance (CD) is reported. If the average ranks of two algorithms are at least CD apart, the difference in the performance of the algorithms is considered statistically significant. The groups of algorithms, for which this does not hold, are connected by red lines.

The interpretation of the results is thus as follows. The top performing group of algorithms, among which there are no statistically significant differences, consists of \textsc{Genie3} and \textsc{URelief}. Moreover, \textsc{Genie3} is statistically significantly better than all the remaining methods (i.e., those outside the top group).
The other two groups are i) \textsc{URelief}, Laplace, \textsc{AgnoS-S}, NDFS, and MCFS, and ii) Laplace, \textsc{AgnoS-S}, NDFS, MCFS, and SPEC.
Thus, the \textsc{Genie3} \emph{method is the new SOTA} in unsupervised feature ranking, whereas \textsc{URelief} is closer in performance to the previous SOTA methods.


\subsection{Parameter Setting Influence on Ranking Performance}\label{sec:hyperparameters}

In the main line of experiments, we set the parameters of our feature ranking methods to the most time-efficient values, where the time complexity critically depends on them.
For ensemble-based scores, we chose extra tree ensembles and $\log_2 n$ as the number of considered features in internal nodes.
As for \textsc{URelief}, we followed the previous findings since the (maximal) number of iterations is not a problem.

In this section, we investigate how the quality of the produced rankings depends on the parameters of the ranking methods. We start with \textsc{URelief} and then proceed to ensemble-based scores.

\begin{figure}
    \centering
    \includegraphics[scale=0.6]{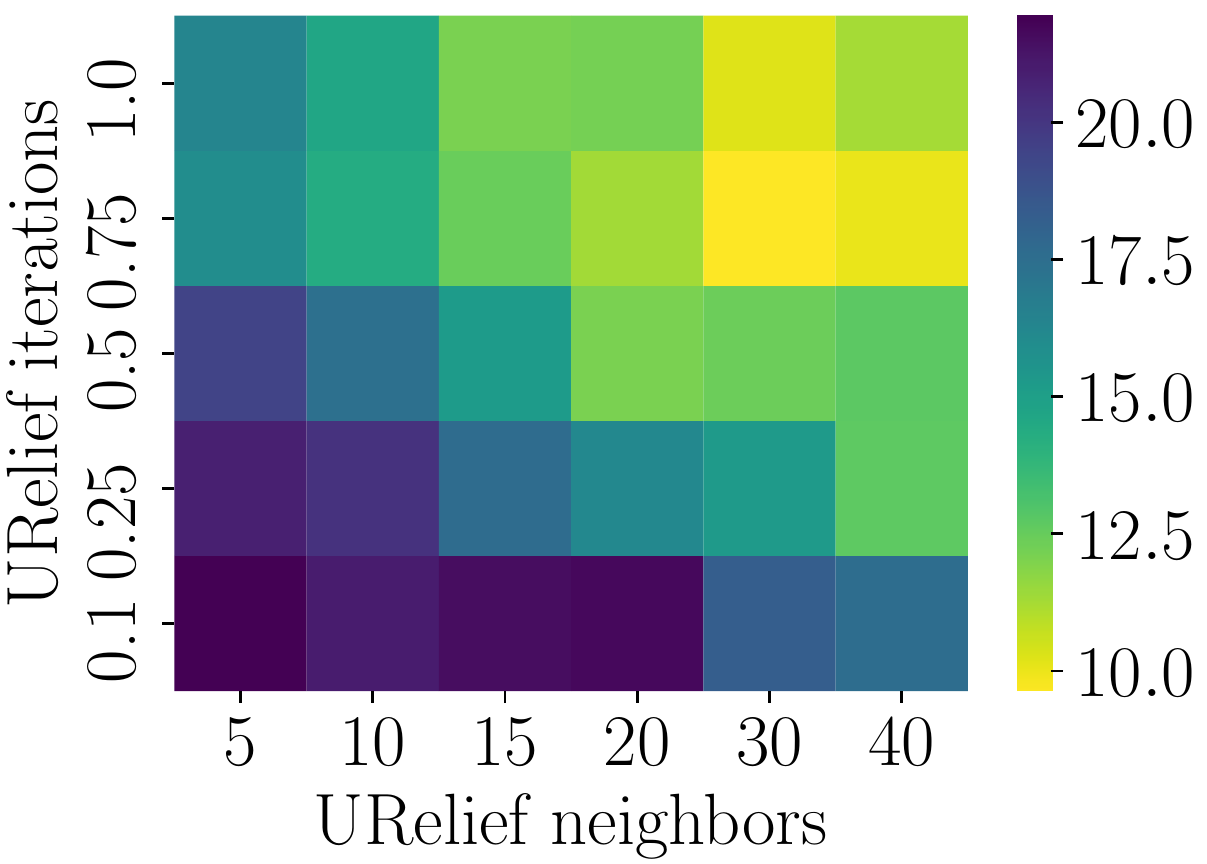}
    \caption{Average ranks (over the considered data sets) of the considered parameter configurations of \textsc{URelief} feature ranking. Lower ranks correspond to better performance.}
    \label{fig:URelief-hyperparameters}
\end{figure}

\textbf{{\normalsize UR}{\scriptsize ELIEF}}. The heat-plot in Figure~\ref{fig:URelief-hyperparameters} shows the average ranks of the feature rankings computed by using all the pairs of different values for the number of iterations $I\in \{0.1m, 0.25m, 0.5m, 0.75m, 1.0m\}$ and the number of neighbors $K\in\{5, 10, 15, 20, 30, 40\}$.
We can see that more neighbors and more iterations in general lead to better performance (lower ranks). Both observations are somewhat expected. More iterations mean more stable estimates of the probabilities present in Eq.~\eqref{eqn:URelief}. 
The explanation for why the algorithm prefers a higher number of neighbors is that the number of features can be really high, thus some noise is definitely present. Therefore, having a higher number of neighbors averages out the noise at least to some extent.
Note that we did not lose much by choosing $I = m$ and $K = 30$, since this is the third-best option.

\begin{figure*}[ht!]
\centering
\begingroup
    \captionsetup[subfigure]{width=0.49\textwidth}
    \subfloat[\textsc{Genie3}\label{fig:ensembles-hyperparameters:genie3}]{
    \includegraphics[scale=0.6]{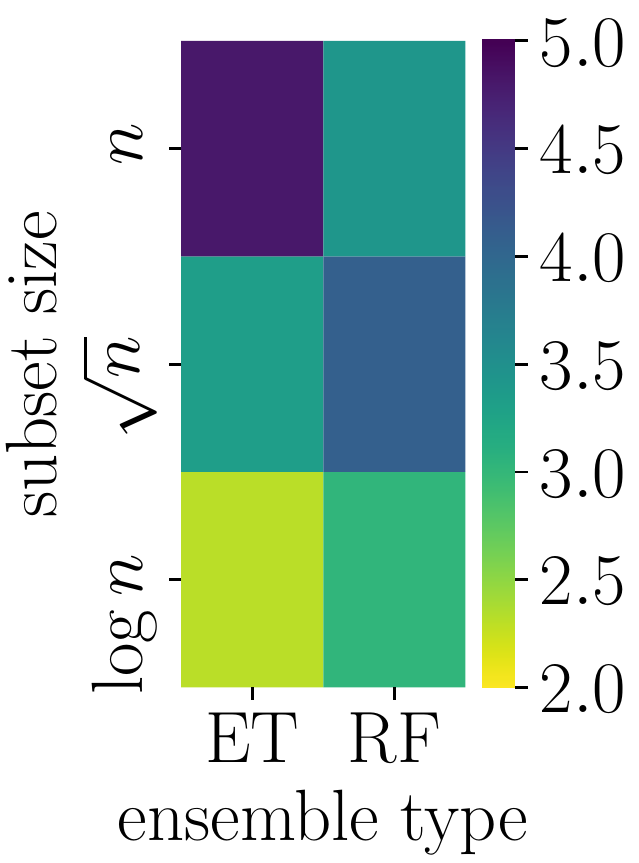}}
\endgroup
\hspace{2cm}
\begingroup
    \centering
    \captionsetup[subfigure]{width=0.45\textwidth}
    \subfloat[\textsc{RandomForest}\label{fig:ensembles-hyperparameters:rf}]{
    \includegraphics[scale=0.6]{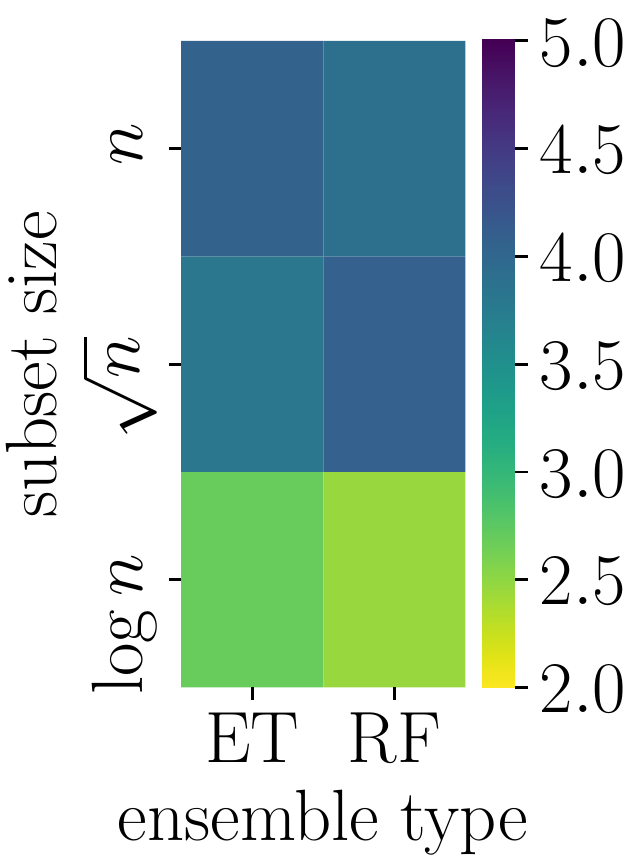}}
\endgroup

\caption{Average ranks (across all data sets) of the considered parameter configurations (three different subset sizes and two ensemble methods), when the feature ranking score is fixed to one of two options (\textsc{Genie3}, \textsc{RandomForest}). Lower ranks correspond to better performance.}
\label{fig:ensembles-hyperparameters}
\end{figure*}

{\bf Ensemble-based scores.} Here, we investigate the influence of the ensemble type (extra trees, random forests, bagging), and the feature subset size. Note that it is enough to consider only extra trees and random forests, since we consider the subset sizes of $n'\in\{\log_2 n, \sqrt{n}, n\}$, and the bagging approach is a special case of random forests when $n' = n$. The heat-plots in Fig.~\ref{fig:ensembles-hyperparameters}
show the performance of the rankings for both considered scores: Genie3 (Fig.~\ref{fig:ensembles-hyperparameters:genie3}) and Random Forest (Fig.~\ref{fig:ensembles-hyperparameters:rf}).

The performance of both scores mostly increases when the subset size $n'$ decreases. A possible explanation for that is that such models overfit less to the data given that the number of features is high. It seems that this parameter is more important that the ensemble method, especially for the \textsc{RandomForest} score.
Again, we did not lose much (performance-wise) when choosing the parameters for the main line of experiments to optimize for time complexity, avoiding the potentially quadratic number of operations in the number of features $n$.

\subsection{Where in a Ranking are the Relevant Features?}\label{sec:curves}
Choosing a number of features to include in a model comes in handy when one wants to evaluate different feature rankings over different domains and many data sets.
However, if a domain expert is interested in a particular data set, a more global view on the ranking might be desirable to understand the problem better. To achieve that, one can build predictive models that use the top 1, 2, 4, \dots, $2^j$, \dots, $2^{\mathit{floor}(\log_2 n)}$, and $n$ features and show the performance of the obtained models as a curve.
Using the geometric (rather than the linear 1, 2, 3,\dots) sequence of numbers of top-ranked features makes the curve construction feasible (building, for example, more than 20 000 predictive models for the GLI-85 data set for the linear sequence might be too time-consuming), while still showing enough details at the beginning of the ranking, which is its most interesting part.

\begin{figure*}[htb!]
\centering
\begingroup
    \captionsetup[subfigure]{width=0.9\textwidth}
    \subfloat[\label{fig:curve:lung}]{
    \includegraphics[scale=0.4]{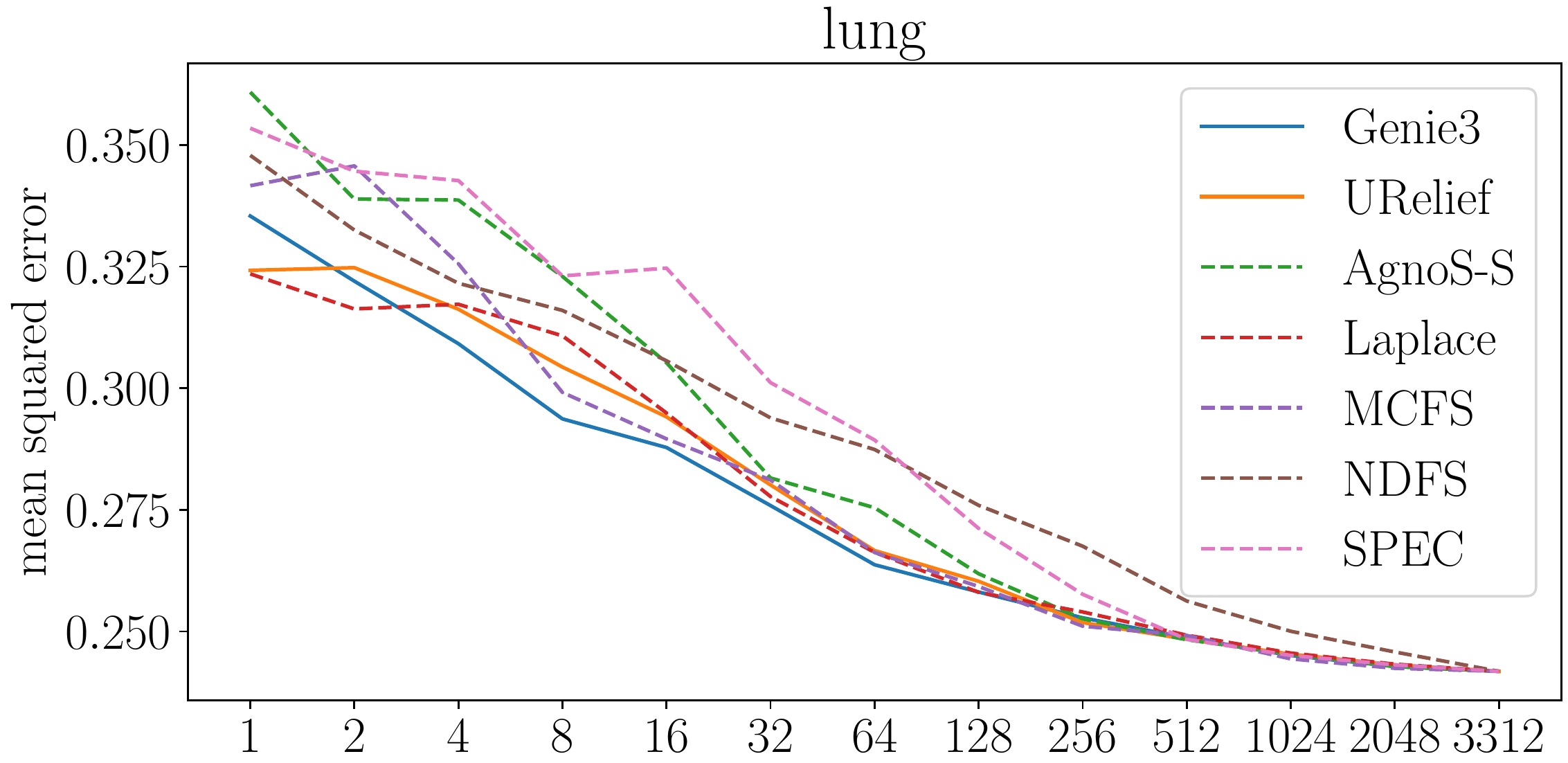}}
\endgroup
\begingroup
    \centering
    \captionsetup[subfigure]{width=0.9\textwidth}
    \subfloat[\label{fig:curve:pcmac}]{
    \includegraphics[scale=0.35]{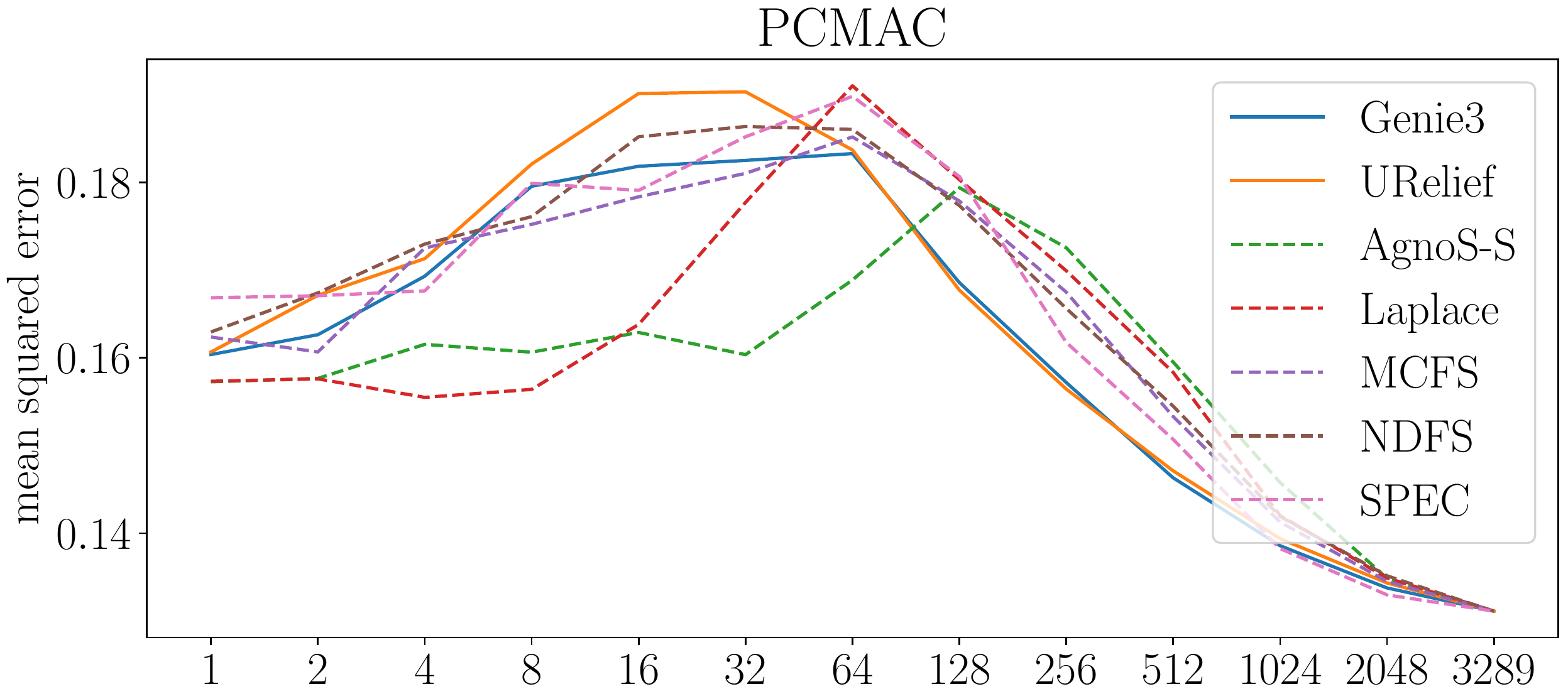}}
\endgroup

\caption{Error curves on two datasets (a) lung and (b) PCMAC for the feature rankings produced by the competing unsupervised ranking methods. Lower placement of a curve corresponds to better performance.
}
\label{fig:curve}
\end{figure*}

Here, we chose two data sets, for which the quality of the rankings is completely different. Note that the results for the rest of the curves are available at \url{https://github.com/Petkomat/unsupervised_ranking}. Figure~\ref{fig:curve:lung} shows the results for the \texttt{lung} data set, where with the increasing number of features, the MSE of the models mostly decreases. This means that, the rankings are mostly correct since relevant features have the highest importance. After \textsc{URelief} and Laplace both discover the most relevant feature, \textsc{Genie3} quickly becomes the one (after four features) that ranks the features best.

The situation in Fig.~\ref{fig:curve:pcmac} is different. These are the rankings for the \texttt{PCMAC} data set. We can see that all the algorithms successfully discover a few most relevant features, but then, mostly useless and noisy features are placed in the ranking after them. This is evident from the increasing MSE. Only after 32 or 64 features, the next relevant ones are positioned and MSE decreases again.

When a domain expert sees those two figures, the expert can locate the position of the relevant features, which will help him understand the problem better. 
Of course, when we use all the features (the last point of every curve), MSE does not depend on the ranking and all the curves end at the same point.

\section{Conclusions}\label{sec:conclusions}
In this work, we propose two novel approaches for unsupervised feature ranking. The first approach uses ensemble-based scores (\textsc{Genie3} and \textsc{RandomForest}), computed from ensembles of predictive clustering trees. The second approach is an unsupervised version of the \textsc{Relief} algorithm (\textsc{URelief}).

After carefully choosing and discussing the evaluation procedure, we conduct an extensive empirical evaluation. We determine how the parameters of both approaches (ensemble-based scores and \textsc{URelief}) influence the quality of the rankings they produce. We show that for the ensemble-based scores, where parameters critically influence the time efficiency, the most efficient rankings fortunately also have the highest quality.

The comparative evaluation shows that the \textsc{Genie3} score works better than the \textsc{RandomForest} score. We then compare \textsc{Genie3} and \textsc{URelief} with five baselines showing that, on average, both proposed methods outperform the baselines. Since the difference in performance between \textsc{Genie3} and the baselines is statistically significant, we recommend, all in all, to use the \textsc{Genie3} score. This score can be efficiently computed from a parallelized ensemble of extremely randomized PCTs, where the feature subset size is set to $n' = \log_2 n$, which typically results in the time complexity of $\mathcal{O}(m n\log m \log n)$, where $m$ and $n$ are the numbers of examples and features.

\section*{Acknowledgements}
The computational experiments presented here were executed on a computing infrastructure from the Slovenian Grid (SLING) initiative, and we thank the administrators Barbara Kra\v{s}ovec and Janez Srakar for their assistance.

\section*{Funding}
This work was supported by the Slovenian Research Agency via the grant P2-0103 and a young researcher grant to MP and B\v{S}. SD and DK also acknowledge the support by the Slovenian Research Agency (via grants J7-9400, J7-1815, J2-9230, and N2-0128), and the European Commission (projects AI4EU (grant number 825619) and TAILOR (grant number 952215)).

\bibliographystyle{acm}
\bibliography{willy}

\end{document}